\documentclass[runningheads]{llncs}
\usepackage{eccv}

\usepackage{eccvabbrv}

\usepackage{graphicx}
\usepackage{booktabs}
\usepackage{svg}
\usepackage{multirow}
\usepackage{array,multirow,graphicx}
\usepackage{float}
\usepackage{stfloats}
\usepackage{soul}

\usepackage[accsupp]{axessibility}  

\usepackage[pagebackref,breaklinks,colorlinks]{hyperref}

\usepackage{orcidlink}

\begin{document}

\title{Evolution of Detection Performance throughout the Online Lifespan of Synthetic Images}

\titlerunning{Evolution of Synthetic Image Detection Performance}

\author{Dimitrios Karageogiou\inst{1} \and
Quentin Bammey\inst{2} \and
Valentin Porcellini\inst{3} \and Bertrand Goupil\inst{3} \and Denis Teyssou\inst{3} \and Symeon Papadopoulos\inst{1}}

\authorrunning{D.~Karageorgiou et al.}

\institute{
    Information Technologies Institute, CERTH, Greece\\\email{\{dkarageo, papadop\}@iti.gr} 
    \and
    Université Paris-Saclay, ENS Paris-Saclay, CNRS, Centre Borelli, France\\ \url{https://bammey.com/}
    \and
    Agence France-Presse, France \\\email{\{valentin.porcellini, bertrand.goupil, denis.teyssou\}@afp.com}
}

\maketitle
\begin{figure}
    \centering
\includesvg[width=0.99\columnwidth]{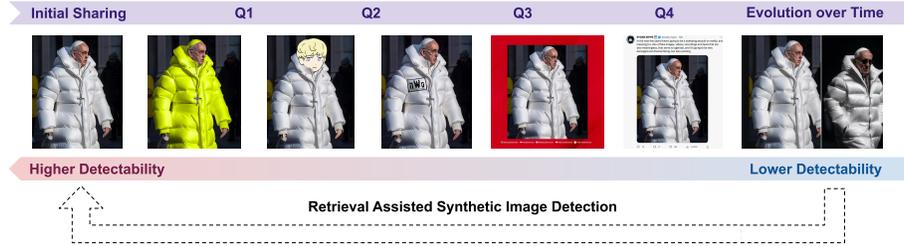}  
    \caption{We study the evolution of synthetic images throughout their online lifespan through collecting different online versions of the ``same'' synthetic image. Using this data, we evaluate state-of-the-art synthetic image detectors, to find out that they mostly fail to detect several instances that were shared online, while the time since initial sharing negatively affects detection performance. Using retrieval-assisted synthetic image detection, it is feasible to retain the initial detection performance throughout the online lifespan of a synthetic image.}
    \label{fig:overview}
\end{figure}

\begin{abstract}
Synthetic images disseminated online significantly differ from those used during the training and evaluation of the state-of-the-art detectors. In this work, we analyze the performance of synthetic image detectors as deceptive synthetic images evolve throughout their online lifespan. Our study reveals that, despite advancements in the field, current state-of-the-art detectors struggle to distinguish between synthetic and real images in the wild. Moreover, we show that the time elapsed since the initial online appearance of a synthetic image negatively affects the performance of most detectors. Ultimately, by employing a retrieval-assisted detection approach, we demonstrate the feasibility to maintain initial detection performance throughout the whole online lifespan of an image and enhance the average detection efficacy across several state-of-the-art detectors by 6.7\% and 7.8\% for balanced accuracy and AUC metrics, respectively. \looseness=-1
\end{abstract}

\section{Introduction}
\label{sec:intro}

Several synthetic image generation approaches have been recently proposed, achieving an outstanding level of photorealism and blending the boundaries between generated and real content \cite{croitoru2023diffusion, zhang2023text}. In response, the image forensics community has explored detecting whether an image originates from a generative model, or constitutes a real one, capturing an actual moment of our physical world \cite{lin2024detecting, akhtar2023deepfakes}. In particular, detectors have been proposed to tackle popular generative approaches, such as Generative Adversarial Networks (GANs) \cite{wang2020cnn, tan2023learning, goodfellow2014generative} and Diffusion Models (DMs) \cite{corvi2023detection, cazenavette2024fakeinversion, ho2020denoising,bammey2023synthbuster,li2024masksim}. All of them achieve very high detection performance in their respective evaluation setups. 
However, when detection approaches are tested on real-life cases of synthetic visual content spreading online, reported user experiences~\cite{dufour2024ammeba} seem to contradict these lab-controlled experimental results. In this work, we analyse how current synthetic image detectors perform when facing synthetic images that circulate online in different variations. Moreover, we study how the evolution of the copies of synthetic images shared online affect the performance of detection methods. 

To evaluate synthetic image detection (SID) approaches on actual cases of AI-generated images spreading online, and to study the evolution of synthetic content with respect to time, suitable benchmark data is required. However, the currently available datasets have either been generated under highly-controlled lab assumptions \cite{wang2020cnn, corvi2023detection, zhu2023genimage, bammey2023synthbuster}, or collected from online channels such as Discord, where images are posted right after their generation \cite{wang2022diffusiondb}. These datasets neither capture the multi-step post-processing operations expected to have been applied on images spreading online, especially for a long time, nor do they consider complex transformations commonly encountered in the wild, such as addition of text and inclusion in memes \cite{dufour2024ammeba}. Benchmarks currently ignore the evolution of a synthetic image throughout its online lifespan. To tackle this limitation and enable a study of the real-world performance of SID methods, we collect the Fact-checked Online Synthetic Image Dataset (FOSID). Starting from some popular fact-checked synthetic images, we collect, curate and analyze a large number of their instances circulating online, thus capturing a web-scale variability of the applied chains of post-processing operations. We capture the dimension of time since the initial online appearance of a synthetic image to enable our study on how the properties of synthetic images, introduced by the generative processes \cite{wang2020cnn, corvi2023detection, wang2023dire}, evolve with respect to the time an image remains online.

We use our FOSID dataset as well as images from several large-scale and forensics-oriented benchmarks to evaluate the performance of several recently proposed SID methods. We find out that all tested methods are improperly calibrated to handle in-the-wild samples. Most of them also fail to produce a well-separated ranking between deceiving synthetic images and real ones originating from the Web. We show that on heavily post-processed instances, such as the ones with additional overlays or included in memes, many detectors plausibly exhibit increased performance by detecting this later post-processing instead of the signal of the synthetic image. In addition, by analyzing the evolution of the synthetic images throughout their online lifespan, we show that the performance of most detectors drops with respect to the time elapsed since the initial online appearance of an image. We exploit this finding by using a Retrieval-Assisted Synthetic Image Detection (RASID) pipeline to maintain consistent SID performance across the different copies of synthetic images, where the longer chain of post-processing operations degrades detection performance. An overview of the proposed concept is presented in \cref{fig:overview}.

Overall, our contributions include the following:
\begin{itemize}
\item We show that the state-of-the-art SID approaches are significantly uncalibrated for handling actual online cases of synthetic images.
\item We find that most SID methods fail to discriminate between synthetic and real image cases collected in the wild with no further preconditions, even when tuning a threshold.
\item We study the effects of chains of post-processing operations applied to an image after its initial online appearance to the performance of SID approaches and notice a degradation over time.
\item We employ a retrieval-assisted detection process to maintain the detection performance across the early- and the late-shared copies of synthetic images, and achieve a performance increase of $6.7\%$ and $7.8\%$ in balanced accuracy and AUC respectively, averaged over several SID methods.
\item We make publicly available the FOSID collection to facilitate further studies on the evolution of fact-checked deceiving synthetic images throughout their online lifespan.
\end{itemize}

\section{Related Work}
\label{sec:related}

\subsection{Synthetic Image Generation}

Throughout the recent years image generation approaches have significantly evolved, rapidly moving from the generation of blurry and low-resolution images~\cite{gregor2015draw} to the creation of photo-realistic imagery indistinguishable from actual photos~\cite{podell2023sdxl}. While several approaches for generative image modelling have been proposed, the progress in the field has been primarily driven by the adoption of the Generative Adversarial Networks (GANs)~\cite{goodfellow2014generative, goodfellow2020generative}, and more recently, by the introduction of the Diffusion Models (DMs)~\cite{ho2020denoising, sohl2015deep}.  

Early unconditional GAN architectures such as ProGAN~\cite{karras2018progressive}, BigGAN~\cite{brock2018large} and StyleGAN~\cite{karras2019style}, pioneered the generation of high-fidelity images using random noise as input. At the same time, conditional GAN architectures, such as CycleGAN~\cite{zhu2017unpaired}, StarGAN~\cite{choi2018stargan}, GauGAN~\cite{park2019semantic} and more recently GigaGAN~\cite{kang2023scaling} introduced the ability to control the image generation process using an image or text input. Lately, the introduction of Diffusion Models has set new standards to generative modelling, with the introduction of architectures such as the denoising diffusion probabilistic models~\cite{ho2020denoising} that learn to reverse the diffusion process and the latent diffusion models~\cite{rombach2022high} that increase efficiency by performing the denoising process in the latent space. Moreover, some recently proposed architectures and methodologies such as ControlNet~\cite{zhang2023adding} and Diffusion in Style~\cite{Everaert_2023_ICCV} have significantly increased the versatility of conditioning pre-trained DMs.     

\subsection{Synthetic Image Detection}

In response to the trends in generative modelling, there has been a surge in methods for detecting synthetic images. Early works in the field primarily focusing on GAN-based architectures noticed that generative models introduce spectral artifacts to the images and proposed methods to model them either in the spatial~\cite{wang2020cnn} or spectral domain~\cite{frank2020leveraging}. Others found that generative models fail to match the distribution of real images in texture-rich regions of the images and proposed approaches for modelling them using either global~\cite{liu2020global} or multi-scale~\cite{ju2022fusing} texture representations. Subsequent works found that the gradients of the latent representations with respect to the RGB images can also discriminate between synthetic and real images~\cite{tan2023learning}.

Recent works have found that DMs also introduce spectral artifacts and adapted previous detectors to the detection of images originating from DMs~\cite{corvi2023detection, bammey2023synthbuster, li2024masksim}. However, the poor generalization performance of such methods due to the differences between the spectral artifacts introduced by generative models even with minimal differences, inspired several works to revisit previously explored ideas, such as the texture artifacts~\cite{zhong2023rich} or the artifacts introduced by the upscaling layers of the generative architectures~\cite{tan2024rethinking}. In pursuit of more generalizable features, recent works exploit errors in the high-level semantics of synthetic images using latent representations from frozen pre-trained encoders, either only in the visual modality~\cite{ojha2023towards, cozzolino2024raising, koutlis2024leveraging} or in the alignment between the visual content and textual captions of the image~\cite{sha2023fake}. In a similar direction, others exploit the errors appearing when reconstructing the real images~\cite{wang2023dire, cazenavette2024fakeinversion}. In addition, recent image forgery localization approaches increase granularity of SID by detecting local manipulations performed by generative models~\cite{guillaro2023trufor, karageorgiou2024fusion, guo2023hierarchical}.

\subsection{Synthetic Image Detection Benchmarks}

Several recent SID works have introduced benchmark datasets spanning a broad range of generative models. Wang et al.~\cite{wang2020cnn} and Asnani et al.~\cite{asnani2023reverse} introduced datasets that include samples from several generative approaches up to the GAN era. More recently, Corvi et al.~\cite{corvi2023detection}, Ojha et al.\cite{ojha2023towards} and Wang et al.~\cite{wang2023dire} have introduced training and testing data generated by diffusion models. Moreover, Synthbuster~\cite{bammey2023synthbuster}, DiffusionDB~\cite{wang2022diffusiondb},  GenImage~\cite{zhu2023genimage} and Twigma~\cite{chen2024twigma} significantly increased the scale of the available data for some of the most popular generative approaches. Last, SIDBench~\cite{schinas2024sidbench} recently introduced a framework for benchmarking SID approaches in a modular manner across different datasets. 

However, all the aforementioned works either employ some automated synthesis pipelines that produce arbitrary synthetic data, or collect synthetic data from online channels where they are posted soon after their generation, including many cartoon-looking and highly styled images. Thus, as Dufour et al.~\cite{dufour2024ammeba} have already highlighted, most of these images are highly unlikely to represent content that would constitute misinformation. Moreover, neither of these datasets capture the evolution of synthetic images over their online lifespan, and as a consequence, they only consider images that have passed through very few (if any) post-processing operations. To the best of our knowledge, we are the first to study the performance of SID approaches in the wild, throughout the online lifespan of synthetic images that are representative of actual cases of misinformation. Thus, we combine i) actual pieces of synthetic imagery constituting misinformation, ii) a representative set of post-processing operations encountered in the wild and iii) the consideration of the time since their online appearance.
\section{Data Collection Process}
\label{sec:dataset}

Assessing the performance of SID methods in the wild requires data that captures the various forms in which a synthetic image may appear online, throughout its online lifespan. We use the term \textit{online lifespan of an image} to denote the evolution of its copies shared online with respect to the time elapsed since its first online appearance. This evolution may include simple post-processing operations, like resizing, cropping or recompression, but also more complex manipulations like the addition of text, the inclusion in memes or the online sharing of screenshots that include the initial image. In this direction, we built a time-ordered collection of the copies of synthetic images that have been previously shared online as pieces of misinformation, namely the Fact-checked Online Synthetic Image Dataset (FOSID).

\begin{figure}[t]
  \centering
  \includesvg[width=1.0\textwidth]{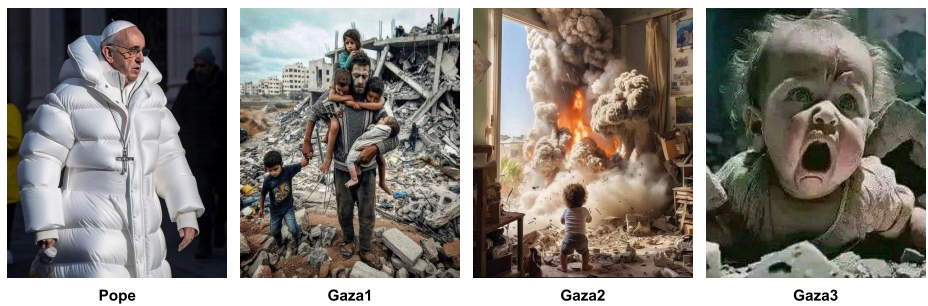}
  \caption{Fact-checked synthetic images used as seeds for the data collection process. The ``Pope'' image is a satirical depiction of the Pope, while the remaining three were presented as relating to events of the Israel-Hamas war. The images have been cropped and scaled to the same aspect ratio for illustration purposes in this figure only.}
  \label{fig:init_images}
\end{figure}

To build this dataset we initially selected a set of four images that have been recently shared online with the intention to deceive and have been proven to originate from synthetic image generators through rigorous fact-checking \footnote{\url{https://factcheck.afp.com/doc.afp.com.33C66F3}} \footnote{\url{https://factcheck.afp.com/doc.afp.com.33ZJ8WU}} \footnote{\url{https://factcheck.afp.com/doc.afp.com.343V9H6}} \footnote{\url{https://www.dailymail.co.uk/news/article-12785455/young-people-Hamas-dirty-tricks-propaganda-war-AMI-H-ORKABY.html}}. These images, that we present in \cref{fig:init_images}, were used as seeds for the collection of four subsets that capture the online evolution of each image with respect to the time since its first online appearance. They were selected as they had seen wide propagation on social media at the time of writing and were also debunked by several fact-checking agencies. To facilitate our data collection process we used the Google Fact Check Tools \cite{googlefctools2024} to perform reverse image search, while at the same time having access to a time-sorted list of the retrieved URLs. In total, we collected $3070$ URLs, spanning an online lifespan between three and 10 months for each of the four seed images, amounting to a total lifespan of 25 months. The evolution over the lifespan of each image constitutes a different subset of FOSID, namely the \textit{Pope, Gaza1, Gaza2} and \textit{Gaza3} subsets.  

\begin{figure}[t]
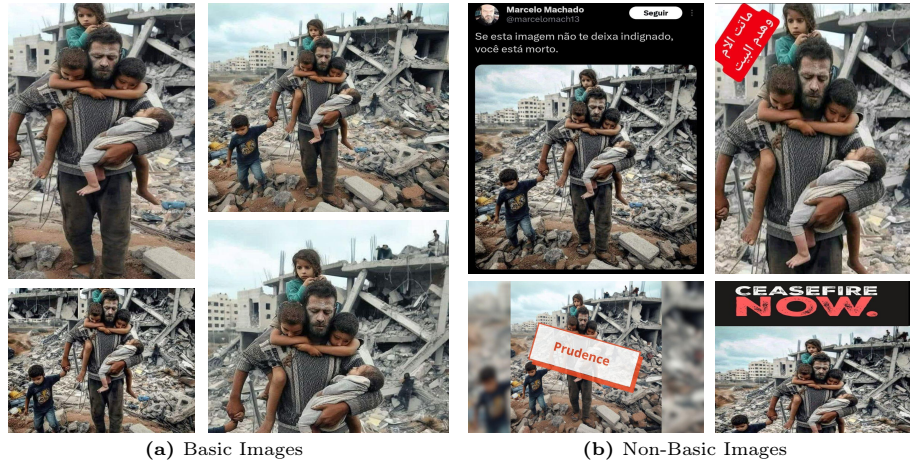

    \centering
    \subfloat[Basic Images]{
        \includesvg[width=0.48\columnwidth]{figures/basic_images_samples.svg}
    } 
    \hfill
    \subfloat[Non-Basic Images]{
        \includesvg[width=0.48\columnwidth]{figures/complex_images_samples.svg}
    }
    \caption{Examples of basic and non-basic images from the Gaza1 subset of FOSID.}
    \label{fig:basic_images_examples}
\end{figure}

The URL list initially returned by the reverse image search process was including irrelevant web content, like URLs of web pages with the seed image under a news feed bar, URLs of web aggregators, and in general URLs of pages where the examined image was not part of the main content. For that reason we manually curated the list of the URLs returned by the reverse image search process, in order to conclude with a list of URLs where each image was the main content. Since many web pages were including more images into their main content along with the image under question, we further extracted the direct URLs pointing to the images visually related to the seed ones. Last, we noticed that a significant number of websites reshare exactly the same image files, without any alterations at the byte-level. We then grouped the URLs according to the unique image files they point to.

\begin{table}
    \centering
    \caption{Overview of the Fact-checked Synthetic Image Dataset (FOSID). The dataset is organized in four subsets, each capturing the evolution of an image over time.}
    \setlength\tabcolsep{0.51em}
\begin{tabular}{l | c | c c c c c}
     \toprule
     \textbf{Subset} & \textbf{Lifespan} & \textbf{Total URLs} & \textbf{Valid URLs} & \textbf{Unique Img} & \textbf{Basic Img} \\
     \midrule
     \textbf{Pope}  & 9 months  & 678 & 664 & 228 & 195 \\
     \textbf{Gaza1} & 3 months  & 772 & 621 & 239 & 146 \\
     \textbf{Gaza2} & 3 months  & 808 & 806 & 272 & 215 \\
     \textbf{Gaza3} & 10 months & 745 & 742 & 216 & 157 \\
     \midrule
     \textbf{Total} & 25 months & 3070 & 2833 & 955 & 713 \\
     \bottomrule
\end{tabular}
    \label{table:dataset_overview}
\end{table}

Several of the collected images had undergone some non-trivial post-processing operations, such as the addition of graphic overlays in the form of text, watermarks or emojis, their inclusion in screenshots, photo collages or memes, or forgery operations, such as background removal. While all the above constitute manipulations that should be expected to be applied to a synthetic image throughout its online lifespan, we recognize that most state-of-the-art SID methods do not explicitly consider them. Thus, we adopt the definition of \textit{basic images} from Dufour et. al \cite{dufour2024ammeba}, for describing images that visually appear to originate from a camera, and cannot be certainly judged otherwise just by visual inspection. Thus, we further curate a \textit{basic} subset of the collected images, which we believe to better represent the handcrafted datasets used for training most state-of-the-art detectors. Also, we consider the basic images to constitute the narrower topic-agnostic range of images a detector should support, since in most cases it is intractable to further distinguish them just by visual inspection and so not realistic to impose such preconditions to the user of the detector. \cref{fig:basic_images_examples} depicts samples belonging to the basic and non-basic subsets of our dataset. The contents of FOSID are summarized in \cref{table:dataset_overview}.  

\section{Evaluation}
\label{sec:evaluation}

\subsection{Metrics}
Initially, we employ the Balanced Accuracy (BA) metric, computed over the $0.5$ threshold, as an indicator of the performance of a detector when deployed in the wild, where it may either be impossible to tune a threshold due to the lack of representative data or the output of the detector is expected to be interpreted as a meaningful probability by the users. Furthermore, we employ the Area Under the ROC Curve (AUC) score as an indicator of the discriminating ability of a detector, and the BA metric, computed over the Equal Error Rate (EER) threshold, to indicate the amount of correctly classified samples of both real and synthetic images, when treating each class as equally important. Moreover, throughout our analysis we found several detectors to perform worse than random prediction, thus, some metrics are reported to be below 50\%. For these cases we intentionally do not inverse the labels to get scores above 50\%, since in the respective evaluation of each detector on lab-generated data an inversion was not considered. This highlights the level of misalignment that is expected to occur when deploying detectors trained on lab-generated data on actual cases. 

\subsection{Synthetic Image Detection Approaches}

In our study we consider several recently proposed SID methods, that are reported to generalize across different generative models in their respective papers. While some of these approaches were originally proposed for GAN-generated images, they all consider cross-model generalization. Thus, we include them to study their generalization performance in the wild, despite the fact that our fact-checked images originate from recent DMs.

\begin{itemize}
    \item \textbf{CNNDetect} \cite{wang2020cnn} that employs a ResNet50 \cite{he2016deep} model  trained on ProGAN data to detect images produced by several GAN-based generative models.
    \item \textbf{FreqDetect} \cite{frank2020leveraging} that uses the DCT transform of an image as input to a classifier to detect GAN-generated images.
    \item \textbf{GramNet} \cite{liu2020global} that leverages global texture representations to detect GAN generated face images.
    \item \textbf{Fusing} \cite{ju2022fusing} that combines global and local features to detect GAN-generated images.
    \item \textbf{LGrad} \cite{tan2023learning} that exploits the gradients of the latent representation of an image produced by an encoder with respect to the input image, to discriminate between GAN-generated and real images.
    \item \textbf{DIMD} \cite{corvi2023detection} that introduces a detection approach based on ResNet50 for detecting images generated by DMs.
    \item \textbf{UnivFD} \cite{ojha2023towards} that uses features from a pre-trained CLIP ViT encoder \cite{radford2021learning} and nearest-neighbor search to detect synthetic images from several generative architectures.
    \item \textbf{DeFake} \cite{sha2023fake} that exploits the difference in the alignment of the image's content and its caption using CLIP's image and text encoders.\cite{radford2021learning}.
    \item \textbf{DIRE} \cite{wang2023dire} that leverages the error in reconstructing an image with a DM to detect fake images.
    \item \textbf{PatchCraft} \cite{zhong2023rich} that exploits the differences in the rich-texture regions between synthetic and real images. 
    \item \textbf{NPR} \cite{tan2024rethinking} that leverages neighboring pixel relationships to capture the artifacts introduced by the upscaling stages of generative models.
    \item \textbf{RINE} \cite{koutlis2024leveraging} that leverages features from the intermediate layers of the CLIP's image encoder to build a general SID model.
\end{itemize}

We used the publicly provided pre-trained models of the methods and evaluated them through the SIDBench~\cite{schinas2024sidbench} framework. For methods that provided more than one pre-trained models, we evaluated all of them and report the results of the best one. In total, we evaluated 25 pre-trained models for the 12 considered detection methods.

\subsection{Sources of Real Images}

In order to capture a web-scale variability of post-processing operations in our real data, we employed images from five public datasets that have either been collected in the wild (before the appearance of high-performing synthetic image generation methods), or have been captured using verified camera images. In particular, we randomly sample 2k images from each of the following datasets:

\begin{itemize}
    \item \textbf{ImageNet} \cite{deng2009imagenet} constituting a diverse collection of highly compressed and post-processed images shared across the Web. 
    \item \textbf{COCO} \cite{lin2014microsoft} as a large-scale source of images focusing on the semantic properties of the depicted objects.
    \item \textbf{Open Images Dataset} \cite{kuznetsova2020open} as a source of high-resolution images that have been shared on the Web.
    \item \textbf{RAISE} \cite{dang2015raise} that includes images captured by professional photographers at the resolution produced by the DSLR camera. Specifically, we use the pre-processed JPEG-RAISE \cite{kwon2022learning} variant, which converts the provided RAW camera files to JPEG ones. This dataset facilitates the evaluation on images without further post-processing, apart from the compression operation.
    \item \textbf{FODB} \cite{hadwiger2021forchheim} that comprises images captured by smartphones and shared over online social media platforms.
\end{itemize}

We draw samples from the validation subsets of ImageNet, COCO and Open Images Dataset to minimize any possible direct or indirect data leaks through the employment of the respective training subsets in the training data of the detectors or their pre-trained backbones. 

\begin{table}[t]
    \centering
    \caption{Evaluation of SID methods using a fixed threshold on the four subsets of FOSID. BA is computed using the 0.5 threshold. The presented scores are averaged over five datasets of real images.}
    \setlength\tabcolsep{0.7em}
\begin{tabular}{ l | c c c c | c }
    \toprule
    \textbf{Approach} & \textbf{Pope} & \textbf{Gaza1} & \textbf{Gaza2} & \textbf{Gaza3} & \textbf{Overall} \\
    \cmidrule(lr){1-1} \cmidrule(lr){2-2} \cmidrule(lr){3-3} \cmidrule(lr){4-4} \cmidrule(lr){5-5} \cmidrule(lr){6-6}
    \textbf{GramNet} \cite{liu2020global}          &43.8 &43.0 &43.0 &43.0 &43.2 \\
    \textbf{UnivFD} \cite{ojha2023towards}         &44.8 &47.7 &45.3 &45.5 &45.8 \\
    \textbf{PatchCraft} \cite{zhong2023rich}       &51.5 &48.3 &48.7 &43.3 &47.9 \\
    \textbf{Fusing} \cite{ju2022fusing}            &49.9 &49.0 &49.0 &49.2 &49.3 \\
    \textbf{CNNDetect} \cite{wang2020cnn}          &49.7 &49.4 &49.4 &49.0 &49.4 \\
    \textbf{FreqDetect} \cite{frank2020leveraging} &57.0 &43.2 &47.6 &50.1 &49.5 \\
    \textbf{Dire} \cite{wang2023dire}              &51.5 &51.5 &51.8 &51.1 &51.5 \\
    \textbf{DIMD} \cite{corvi2023detection}        &70.6 &\textbf{78.0} &51.7 &52.8 &53.0 \\
    \textbf{NPR} \cite{tan2024rethinking}          &\ul{74.0} &74.0 &26.8 &\ul{73.6} &59.2 \\
    \textbf{Rine} \cite{koutlis2024leveraging}     &46.4 &82.1 &\ul{67.8} &51.0 &61.8 \\
    \textbf{LGrad} \cite{tan2023learning}          &61.0 &58.9 &\textbf{71.2} &\textbf{81.5} &\ul{68.1} \\
    \textbf{DeFake} \cite{sha2023fake}             &\textbf{81.1} &\ul{72.9} &64.6 &72.6 &\textbf{72.8} \\
    \bottomrule
\end{tabular}
    \label{table:calibration_eval}
\end{table}

\subsection{Evaluation of synthetic image detectors using a fixed threshold}

When deploying SID methods in the wild where they may encounter images from unknown generative models with arbitrary post-processing operations applied to them, calibrating the detection threshold can be infeasible. To provide a meaningful detection score to a user who might interpret it as the probability of an image being synthetic, the scores for any synthetic images should lie close to $1$ while the scores for any real ones close to $0$. Thus, we evaluate all the detection methods by computing the BA metric using the common threshold of $0.5$ and we report it for each FOSID subset in \cref{table:calibration_eval}. The performance of the majority of methods is very close to random selection, while the highest performing achieves a BA value of $72.8\%$, which is significantly lower than its performance on lab-collected data \cite{sha2023fake}. Thus, most models appear to be performing much worse in the wild. 

\subsection{Evaluation of the discriminatory ability of detectors}

\begin{table}[t]
    \centering
    \caption{Evaluation of the discrimination capability of SID methods on the four subsets of FOSID. BA computed using the EER threshold and the AUC metrics are reported. The presented scores are averaged over five datasets of real images.}
    \setlength\tabcolsep{0.34em}
\begin{tabular}{ l | cc cc cc cc | cc }
    \toprule
    \multirow{2}{*}{\textbf{Approach}} &\multicolumn{2}{c}{\textbf{Pope}} &\multicolumn{2}{c}{\textbf{Gaza1}} &\multicolumn{2}{c}{\textbf{Gaza2}} &\multicolumn{2}{c}{\textbf{Gaza3}} &\multicolumn{2}{c}{\textbf{Overall}} \\
    &ACC &AUC &ACC &AUC &ACC &AUC &ACC &AUC &ACC &AUC \\
    \cmidrule(lr){1-1} \cmidrule(lr){2-3} \cmidrule(lr){4-5} \cmidrule(lr){6-7} \cmidrule(lr){8-9} \cmidrule(lr){10-11}
    \textbf{GramNet} \cite{liu2020global}            &33.2 &30.7 &28.0 &21.6 &21.7 &12.0 &17.4 &11.5 &25.1 &19.0 \\
    \textbf{UnivFD} \cite{ojha2023towards}           &31.8 &28.9 &45.5 &49.9 &54.2 &57.5 &33.8 &34.3 &41.3 &42.6 \\
    \textbf{FreqDetect} \cite{frank2020leveraging}   &57.8 &59.5 &38.1 &37.0 &48.5 &49.9 &55.2 &56.1 &49.9 &50.6 \\
    \textbf{PatchCraft} \cite{zhong2023rich}         &59.7 &62.2 &55.0 &56.7 &53.9 &56.2 &45.5 &44.5 &53.5 &54.9 \\
    \textbf{NPR} \cite{tan2024rethinking}            &58.5 &58.4 &80.0 &80.8 &37.3 &36.5 &55.9 &55.8 &57.9 &57.9 \\
    \textbf{Fusing} \cite{ju2022fusing}              &59.6 &61.7 &57.5 &57.5 &57.0 &61.3 &64.6 &68.6 &59.7 &62.3 \\
    \textbf{CNNDetect} \cite{wang2020cnn}            &68.3 &73.5 &55.8 &57.7 &61.6 &70.8 &50.0 &49.2 &58.9 &62.8 \\
    \textbf{Dire} \cite{wang2023dire}                &50.0 &63.8 &50.0 &65.6 &50.6 &68.3 &50.0 &64.1 &50.1 &65.5 \\
    \textbf{LGrad} \cite{tan2023learning}            &65.2 &67.5 &59.9 &63.4 &\ul{71.1} &79.0 &\textbf{84.0} &\textbf{90.0} &70.1 &75.0 \\
    \textbf{Rine} \cite{koutlis2024leveraging}       &51.5 &51.7 &\textbf{84.6} &\textbf{90.0} &\textbf{79.0} &\textbf{88.1} &69.9 &74.0 &71.2 &75.9 \\
    \textbf{DeFake} \cite{sha2023fake}               &\textbf{84.1} &\textbf{91.3} &72.7 &79.4 &64.2 &69.9 &\ul{71.7} &\ul{77.0} &\ul{73.2} &\ul{79.4} \\
    \textbf{DIMD} \cite{corvi2023detection}          &\ul{81.3} &\ul{89.0} &\ul{81.9} &\ul{89.6} &70.1 &\ul{80.6} &66.6 &71.8 &\textbf{75.0} &\textbf{82.8} \\
    \bottomrule
\end{tabular}

    \label{table:overall_eval}
\end{table}

In order to answer whether the state-of-the-art detectors can correctly discriminate synthetic images from real, irrespective of the detection threshold, we compute the BA metric using the EER threshold of each detector and the threshold-agnostic AUC score. We report the results on \cref{table:overall_eval} and note that while the accuracy of the best performing model in the list has slightly improved to $75\%$ compared to the performance achieved using a fixed threshold in \cref{table:calibration_eval}, it is still far below the performance reported by the same method on lab-generated data \cite{corvi2023detection}. Furthermore, we notice that two detectors inversely predict the true labels, thus performing significantly worse than random selection. Overall, the detection performance of most detectors indicates that there is still significant progress needed until it becomes feasible to separate synthetic from real images in the wild. 

\subsection{Evaluation of detectors on different sources of real data}

\begin{table}[t]
    \centering
    \caption{Evaluation of the discrimination capability of SID methods on five real datasets. BA computed using the EER threshold and the AUC metrics are reported. The presented scores are averaged over the four subsets of FODB.}
    \setlength\tabcolsep{0.19em}
\begin{tabular}{l | cc cc cc cc cc | cc}
\toprule
\multirow{2}{*}{\textbf{Approach}} &\multicolumn{2}{c}{\textbf{COCO}} &\multicolumn{2}{c}{\textbf{ImageNet}} &\multicolumn{2}{c}{\textbf{OpenIm.}} &\multicolumn{2}{c}{\textbf{RAISE}} &\multicolumn{2}{c}{\textbf{FODB}} &\multicolumn{2}{c}{\textbf{Overall}} \\
&ACC &AUC &ACC &AUC &ACC &AUC &ACC &AUC &ACC &AUC &ACC &AUC \\
\cmidrule{1-1} \cmidrule{2-3} \cmidrule{4-5} \cmidrule{6-7} \cmidrule{8-9} \cmidrule{10-11} \cmidrule{12-13}
\textbf{GramNet} &19.7 &13.5 &14.7 &9.3 &26.1 &14.2 &28.4 &23.8 &36.6 &34.1 &25.1 &19.0 \\
\textbf{UnivFD} &50.7 &53.5 &55.4 &59.8 &36.5 &39.3 &23.5 &18.8 &40.6 &42.0 &41.3 &42.6 \\
\textbf{FreqDetect} &47.3 &46.3 &59.0 &60.8 &48.7 &50.0 &46.1 &46.8 &48.4 &49.4 &49.9 &50.6 \\
\textbf{RPTC} &56.0 &58.0 &73.7 &79.5 &49.4 &51.2 &44.7 &43.0 &43.6 &42.8 &53.5 &54.9 \\
\textbf{NPR} &19.9 &19.0 &\textbf{96.4} &\textbf{98.5} &66.3 &63.2 &13.3 &13.0 &\textbf{93.5} &\textbf{95.7} &57.9 &57.9 \\
\textbf{Fusing} &73.5 &78.8 &69.1 &73.0 &56.4 &61.3 &44.6 &41.8 &54.8 &56.7 &59.7 &62.3 \\
\textbf{CNNDetect} &70.3 &76.8 &68.3 &74.3 &56.2 &64.0 &45.0 &42.0 &54.8 &57.0 &58.9 &62.8 \\
\textbf{Dire} &50.0 &62.7 &50.0 &66.0 &50.7 &68.3 &50.0 &\ul{69.5} &50.0 &60.8 &50.1 &65.5 \\
\textbf{LGrad} &80.3 &86.8 &70.0 &75.5 &70.4 &75.3 &\ul{61.2} &60.0 &68.4 &73.5 &70.1 &75.0 \\
\textbf{Rine} &\textbf{83.7} &\ul{89.0} &86.7 &92.3 &72.0 &\ul{81.0} &49.1 &50.2 &64.7 &67.2 &71.2 &75.9 \\
\textbf{DeFake} &76.3 &83.0 &71.1 &77.0 &\textbf{73.5} &80.0 &\textbf{74.0} &\textbf{80.0} &71.3 &77.0 &\ul{73.2} &\ul{79.4} \\
\textbf{DIMD} &\ul{82.5} &\textbf{90.0} &\ul{88.0} &\ul{95.0} &\ul{69.2} &\textbf{82.0} &61.1 &64.8 &\ul{74.0} &\ul{82.1} &\textbf{75.0} &\textbf{82.8} \\
\cmidrule{1-1} \cmidrule{2-3} \cmidrule{4-5} \cmidrule{6-7} \cmidrule{8-9} \cmidrule{10-11} \cmidrule{12-13}
\textbf{Overall} &59.2 &63.1 &66.9 &71.7 &56.3 &60.8 &45.1 &46.1 &58.4 &61.5 &57.2 &60.7 \\
\bottomrule
\end{tabular}
    \label{table:real_data_eval}
\end{table}

\begin{table}[t]
    \centering
    \caption{Evaluation of SID methods across all images of FOSID and only its basic subset. The BA and the AUC metrics are reported. The presented scores are averaged over the four main subsets of FOSID and over five datasets of real images.}
    \setlength\tabcolsep{0.6em}
\definecolor{negc}{HTML}{8f003a}
\definecolor{posc}{HTML}{008f3a}
\newcommand{\negv}[1]{\textcolor{negc}{#1}}
\newcommand{\posv}[1]{\textcolor{posc}{#1}}
\begin{tabular}{ l | cc cc | cc }
    \toprule
    \multirow{2}{*}{\textbf{Approach}} & \multicolumn{2}{c}{\textbf{All}} & \multicolumn{2}{c}{\textbf{Basic}} & \multicolumn{2}{c}{\textbf{Diff. (\%)}} \\
    & ACC & AUC & ACC & AUC  & ACC & AUC \\     
    \cmidrule(lr){1-1} \cmidrule(lr){2-3} \cmidrule(lr){4-5} \cmidrule(lr){6-7}
    \textbf{GramNet} \cite{liu2020global}          &25.1 &19.0 &20.3 &15.6 &\negv{-19.2} &\negv{-18.0} \\
    \textbf{UnivFD} \cite{ojha2023towards}         &41.3 &42.6 &37.4 &37.0 &\negv{-9.5} &\negv{-13.2} \\
    \textbf{NPR} \cite{tan2024rethinking}          &57.9 &57.9 &41.6 &41.9 &\negv{-28.1} &\negv{-27.6} \\
    \textbf{FreqDetect} \cite{frank2020leveraging} &49.9 &50.6 &50.0 &51.0 &\posv{0.2} &\posv{0.7} \\
    \textbf{Rine} \cite{koutlis2024leveraging}     &53.5 &54.9 &54.9 &56.0 &\posv{2.6} &\posv{2.0} \\
    \textbf{CNNDetect} \cite{wang2020cnn}          &58.9 &62.8 &56.7 &58.5 &\negv{-3.8} &\negv{-6.9} \\
    \textbf{Fusing} \cite{ju2022fusing}            &59.7 &62.3 &57.6 &58.6 &\negv{-3.5} &\negv{-6.0} \\
    \textbf{Dire} \cite{wang2023dire}              &50.1 &65.5 &50.0 &68.6 &\negv{-0.3} &\posv{4.8} \\
    \textbf{PatchCraft} \cite{zhong2023rich}       &71.2 &75.9 &69.8 &73.7 &\negv{-2.0} &\negv{-3.0} \\
    \textbf{LGrad} \cite{tan2023learning}          &70.1 &75.0 &70.1 &74.2 &\posv{0.1} &\negv{-1.0} \\
    \textbf{DIMD} \cite{corvi2023detection}        &\textbf{75.0} &\textbf{82.8} &\ul{73.6} &\ul{79.1} &\negv{-1.8} &\negv{-4.5} \\
    \textbf{DeFake} \cite{sha2023fake}             &\ul{73.2} &\ul{79.4} &\textbf{73.7} &\textbf{80.4} &\posv{0.7} &\posv{1.3} \\
    \midrule
    \multicolumn{4}{c}{}                           &                     \textbf{Overall} & \negv{\textbf{-5.4}} & \negv{\textbf{-6.0}} \\
    
\end{tabular}
    \label{table:basic_eval}
\end{table}

To study how the SID methods perform with respect to different sources of real data, we report in \cref{table:real_data_eval} the performance of several detectors when considering separately each of the five datasets of real images. We present the BA using the EER threshold and the AUC metrics. The results show that most detectors are better aligned to the ImageNet data, achieving on average $66.9\%$ on balanced accuracy and $71.7\%$ on AUC across all detectors. Instead, the most challenging source of real images is the RAISE dataset, with $45.1\%$ and $46.1\%$ on the same metrics respectively. While it may seem as counter-intuitive that the forensics-oriented JPEG-RAISE dataset, which involves minimal post-processing compared to the rest, to be the most challenging one, a reason for its challenging nature is likely that it includes several megapixel images, something that most detectors do not typically consider in their training. Furthermore, we note that the easiest data from the validation splits of COCO and ImageNet, align very closely with the validation and testing data of several detectors, that consider them. 

\begin{table}[t]
    \centering
    \caption{Evaluation over the online lifespan of images. BA computed using the EER threshold and the AUC metrics are reported. The presented scores are averaged over the four main subsets of FOSID and over five datasets of real images.}
    \setlength\tabcolsep{0.7em}
\definecolor{negc}{HTML}{8f003a}
\definecolor{posc}{HTML}{008f3a}
\newcommand{\negv}[1]{\textcolor{negc}{#1}}
\newcommand{\posv}[1]{\textcolor{posc}{#1}}
\begin{tabular}{l cc cc | cc}
\toprule
\multirow{2}{*}{\textbf{Approach}} &\multicolumn{2}{c}{\textbf{Q1}} &\multicolumn{2}{c}{\textbf{Q4}} &\multicolumn{2}{c}{\textbf{Diff. (\%)}} \\\cmidrule{2-7}
&ACC &AUC &ACC &AUC &ACC &AUC  \\
\cmidrule(lr){1-1} \cmidrule(lr){2-3} \cmidrule(lr){4-5} \cmidrule(lr){6-7}
\textbf{GramNet} \cite{liu2020global} &21.5 &17.6 &21.3 &17.4 &\negv{-0.9} &\negv{-1.1} \\
\textbf{UnivFD} \cite{ojha2023towards} &42.4 &42.6 &41.4 &42.2 &\negv{-2.5} &\negv{-0.9} \\
\textbf{FreqDetect} \cite{frank2020leveraging} &55.1 &56.5 &48.2 &47.9 &\negv{-12.5} &\negv{-15.2} \\
\textbf{PatchCraft} \cite{zhong2023rich}  &56.0 &57.3 &53.0 &54.5 &\negv{-5.3} &\negv{-4.9} \\
\textbf{NPR} \cite{tan2024rethinking} &57.6 &58.7 &57.8 &57.8 &\posv{0.2} &\negv{-1.6} \\
\textbf{Fusing} \cite{ju2022fusing} &59.0 &61.3 &58.2 &59.9 &\negv{-1.3} &\negv{-2.3} \\
\textbf{CNNDetect} \cite{wang2020cnn} &60.0 &62.9 &56.5 &59.3 &\negv{-5.9} &\negv{-5.7} \\
\textbf{Dire} \cite{wang2023dire} &50.0 &68.3 &50.0 &65.4 &0.0 &\negv{-4.2} \\
\textbf{LGrad} \cite{tan2023learning} &70.4 &74.8 &72.1 &76.7 &\posv{2.5} &\posv{2.6} \\
\textbf{Rine} \cite{koutlis2024leveraging} &71.8 &76.0 &71.0 &75.3 &\negv{-1.2} &\negv{-0.9} \\
\textbf{DeFake} \cite{sha2023fake} &72.7 &78.8 &72.9 &79.9 &\posv{0.2} &\posv{1.4} \\
\textbf{DIMD} \cite{corvi2023detection} &78.2 &84.4 &74.3 &80.2 &\negv{-5.0} &\negv{-5.0} \\
\midrule
\multicolumn{4}{c}{} & \textbf{Overall} &\negv{\textbf{-2.6}} &\negv{\textbf{-3.2}} \\
\end{tabular}
    \label{table:time_eval}
\end{table}

\begin{table}
    \centering
    \caption{Evaluation of RASID. BA computed using the EER threshold and the AUC metrics are reported. The presented scores are averaged over the four main subsets of FOSID and over five datasets of real images.}
    \setlength\tabcolsep{0.7em}
\definecolor{negc}{HTML}{8f003a}
\definecolor{posc}{HTML}{008f3a}
\newcommand{\negv}[1]{\textcolor{negc}{#1}}
\newcommand{\posv}[1]{\textcolor{posc}{#1}}
\begin{tabular}{l cc cc | cc}
\toprule
\multirow{2}{*}{\textbf{Approach}} &\multicolumn{2}{c}{\textbf{Direct Det.}} &\multicolumn{2}{c}{\textbf{RASID}} &\multicolumn{2}{c}{\textbf{Diff. (\%)}} \\
\cmidrule{2-7}
&ACC &AUC &ACC &AUC &ACC &AUC \\
\cmidrule(lr){1-1} \cmidrule(lr){2-3} \cmidrule(lr){4-5} \cmidrule(lr){6-7}
\textbf{GramNet} \cite{liu2020global} &20.3 &15.6 &21.8 &17.5 &\posv{7.0} &\posv{13.0} \\
\textbf{UnivFD} \cite{ojha2023towards} &37.4 &37.0 &41.0 &41.7 &\posv{10.0} &\posv{13.0} \\
\textbf{FreqDetect} \cite{frank2020leveraging} &50.0 &51.0 &52.2 &53.6 &\posv{4.0} &\posv{5.0} \\
\textbf{PatchCraft} \cite{zhong2023rich} &54.9 &56.0 &55.6 &56.6 &\posv{1.0} &\posv{1.0} \\
\textbf{NPR} \cite{tan2024rethinking} &41.6 &41.9 &57.5 &58.4 &\posv{38.0} &\posv{39.0} \\
\textbf{Fusing} \cite{ju2022fusing} &57.6 &58.6 &60.2 &61.9 &\posv{4.0} &\posv{6.0} \\
\textbf{CNNDetect} \cite{wang2020cnn} &56.7 &58.5 &59.3 &62.3 &\posv{5.0} &\posv{7.0} \\
\textbf{Dire} \cite{wang2023dire} &50.0 &68.6 &50.0 &67.6 &0.0 &\negv{-2.0} \\
\textbf{LGrad} \cite{tan2023learning} &70.1 &74.2 &71.3 &75.7 &\posv{2.0} &\posv{2.0} \\
\textbf{Rine} \cite{koutlis2024leveraging} &69.8 &73.7 &\ul{73.2} &77.1 &\posv{5.0} &\posv{5.0} \\
\textbf{DeFake} \cite{sha2023fake} &\textbf{73.7} &\textbf{80.4} &72.9 &\ul{79.5} &\negv{-1.0} &\negv{-1.0} \\
\textbf{DIMD} \cite{corvi2023detection} &\ul{73.6} &\ul{79.1} &\textbf{77.6} &\textbf{83.6} &\posv{5.0} &\posv{6.0} \\
\midrule
\multicolumn{4}{c}{} & \textbf{Overall} &\posv{\textbf{6.7}} &\posv{\textbf{7.8}} \\
\end{tabular}
    \label{table:retrieval_assisted_performance}
\end{table}

\subsection{Evaluation on basic images}

Most state-of-the-art SID methods primarily consider basic images for their training. Thus, in this experiment we evaluate their performance by considering only the subset of basic images in FOSID and report the BA computed over the EER threshold and the AUC score in \cref{table:basic_eval}, while we sort models based on their AUC score on basic images. We show that in eight out of 12 considered models there is a performance drop, ranging in relative values from $-0.3\%$ to $-28.1\%$ in the case of BA and from $-1\%$ to $-27.6\%$ in the case of AUC. Instead, the gains for the four remaining models range from $0.1\%$ to $2.6\%$ for BA and from $0.7\%$ to $4.8\%$ for AUC. Overall, there is an average performance drop across all detectors of $-5.4\%$ and $-6.0\%$, and a drop of $-1.7\%$ and $-2.9\%$ between the best detector on all and only the basic samples, for the metrics of BA and AUC respectively. We attribute this drop to the fact that many detectors are sensitive to the non-trivial post-processing operations applied to the non-basic images, and thus detect these later manipulations, instead of the primary synthetic image signal.


\subsection{Detection performance throughout the online lifespan of images}

In this experiment we attempt to answer how the evolution of the copies of a synthetic image with respect to the time elapsed since its first online appearance affects SID performance. To this end, we split each of the time-sorted subsets of FOSID  originating from a different seed image into quarters and measure the performance difference between the first (Q1) and fourth quarter (Q4) images. We use the basic images to align as much as possible with the training data of the detectors. We report the BA computed over the EER threshold and the AUC metrics in \cref{table:time_eval}. In the majority of  detectors there is a relative performance drop ranging between $-0.9\%$ and $-15.2\%$ and between $-0.9\%$ and $-12.5\%$ for the AUC and BA metrics respectively. Overall, we note an average performance drop of $-3.2\%$ and $-2.6\%$ in the same metrics. 9 of the 12 methods present a drop in ACC (including one which was already at random-like accuracy), and 10 of them a drop in AUC. Thus, we see that the accumulation of post-processing operations into the subsequent copies of a synthetic image shared online, degrade the ability of the state-of-the-art detectors to distinguish them from real images, even though those post-processing operations are mostly invisible in the case of basic images that we examine.
\vspace{-12pt}

\subsection{Retrieval-assisted detection of synthetic images}

To recover the performance drop that occurs as the time since the initial online appearance of a synthetic image increases, we introduce the concept of Retrieval Assisted Synthetic Image Detection (RASID). In particular, we employ the DnS \cite{kordopatis2022dns} architecture, proposed for near-duplicate media retrieval, to create an index of all images submitted to the detection system. Then, for images submitted to the system after the Q1 range, instead of directly using the scores produced by the detectors, we query the retrieval system with the submitted image to return all near-duplicate images submitted throughout Q1. If such images exist in the index, the mean of their detection scores with respect to each detection approach is used. Otherwise, the direct output of the detector is used. The empirically selected similarity threshold of $0.7$ is used to classify the images retrieved by DnS as near-duplicates of the query one.  

To better align with the training data of the detectors, we evaluate the RASID approach using the basic images of FOSID. In \cref{table:retrieval_assisted_performance} we report the BA computed over the EER threshold and the AUC metrics for both the cases when directly considering the output of a detector and when applying the RASID approach. We show that the later improves SID performance across all the detectors by $6.7\%$ and $7.8\%$ on the BA and AUC metrics respectively.   

\section{Conclusion}
\label{sec:conclusion}

We analyzed for the first time the performance of SID approaches on actual online misinformation cases of synthetic images. In our study, we found that the current state-of-the-art detectors fail to discriminate between real and synthetic images that circulate online, thus indicating that the unconditional in the wild detection of deceiving synthetic images is not yet reliably addressed. Furthermore, we found that neither the calibration of a threshold, nor the visual alignment of online images with the ones considered during the training of the detectors is sufficient to improve SID performance. This highlights the need for further exploration in the direction of capturing artifacts that are robust to deep chains of post-processing operations that are commonly encountered in online images. Furthermore, we presented that as the time since the initial sharing of a synthetic image passes, the accumulation of post-processing operations degrades the ability to detect its subsequent copies. Using a retrieval-assisted detection pipeline, we showed that it is feasible to maintain consistent SID performance for the entire online lifespan of an image. This indicates that the effective combination of SID and image retrieval approaches constitutes a promising research direction, since the later primarily rely on image content, and thus, are expected to be less susceptible to complex post-processing operations.

\medskip\noindent\textbf{Acknowledgments:}
This work has received funding by the European Union under the Horizon Europe vera.ai project, grant agreement number 101070093. It has also received funding by the ANR under the APATE project, grant number ANR-22-CE39-0016.\\Centre Borelli is also a member of Université Paris Cité, SSA and INSERM. 

\clearpage

%
%
\bibliographystyle{splncs04}
\bibliography{main}

\begin{thebibliography}{10}
\providecommand{\url}[1]{\texttt{#1}}
\providecommand{\urlprefix}{URL }
\providecommand{\doi}[1]{https://doi.org/#1}

\bibitem{akhtar2023deepfakes}
Akhtar, Z.: Deepfakes generation and detection: a short survey. Journal of Imaging  \textbf{9}(1), ~18 (2023)

\bibitem{asnani2023reverse}
Asnani, V., Yin, X., Hassner, T., Liu, X.: Reverse engineering of generative models: Inferring model hyperparameters from generated images. IEEE Transactions on Pattern Analysis and Machine Intelligence  (2023)

\bibitem{bammey2023synthbuster}
Bammey, Q.: Synthbuster: Towards detection of diffusion model generated images. IEEE Open Journal of Signal Processing  (2023)

\bibitem{brock2018large}
Brock, A., Donahue, J., Simonyan, K.: Large scale gan training for high fidelity natural image synthesis. In: International Conference on Learning Representations (2018)

\bibitem{cazenavette2024fakeinversion}
Cazenavette, G., Sud, A., Leung, T., Usman, B.: Fakeinversion: Learning to detect images from unseen text-to-image models by inverting stable diffusion. In: Proceedings of the IEEE/CVF Conference on Computer Vision and Pattern Recognition. pp. 10759--10769 (2024)

\bibitem{chen2024twigma}
Chen, Y., Zou, J.Y.: Twigma: A dataset of ai-generated images with metadata from twitter. Advances in Neural Information Processing Systems  \textbf{36} (2024)

\bibitem{choi2018stargan}
Choi, Y., Choi, M., Kim, M., Ha, J.W., Kim, S., Choo, J.: Stargan: Unified generative adversarial networks for multi-domain image-to-image translation. In: Proceedings of the IEEE conference on computer vision and pattern recognition. pp. 8789--8797 (2018)

\bibitem{corvi2023detection}
Corvi, R., Cozzolino, D., Zingarini, G., Poggi, G., Nagano, K., Verdoliva, L.: On the detection of synthetic images generated by diffusion models. In: ICASSP 2023-2023 IEEE International Conference on Acoustics, Speech and Signal Processing (ICASSP). pp.~1--5. IEEE (2023)

\bibitem{cozzolino2024raising}
Cozzolino, D., Poggi, G., Corvi, R., Nie{\ss}ner, M., Verdoliva, L.: Raising the bar of ai-generated image detection with clip. In: Proceedings of the IEEE/CVF Conference on Computer Vision and Pattern Recognition. pp. 4356--4366 (2024)

\bibitem{croitoru2023diffusion}
Croitoru, F.A., Hondru, V., Ionescu, R.T., Shah, M.: Diffusion models in vision: A survey. IEEE Transactions on Pattern Analysis and Machine Intelligence  \textbf{45}(9),  10850--10869 (2023)

\bibitem{dang2015raise}
Dang-Nguyen, D.T., Pasquini, C., Conotter, V., Boato, G.: Raise: A raw images dataset for digital image forensics. In: Proceedings of the 6th ACM multimedia systems conference. pp. 219--224 (2015)

\bibitem{deng2009imagenet}
Deng, J., Dong, W., Socher, R., Li, L.J., Li, K., Fei-Fei, L.: Imagenet: A large-scale hierarchical image database. In: 2009 IEEE conference on computer vision and pattern recognition. pp. 248--255. Ieee (2009)

\bibitem{dufour2024ammeba}
Dufour, N., Pathak, A., Samangouei, P., Hariri, N., Deshetti, S., Dudfield, A., Guess, C., Escayola, P.H., Tran, B., Babakar, M., et~al.: Ammeba: A large-scale survey and dataset of media-based misinformation in-the-wild. arXiv preprint arXiv:2405.11697  (2024)

\bibitem{Everaert_2023_ICCV}
Everaert, M.N., Bocchio, M., Arpa, S., S\"usstrunk, S., Achanta, R.: Diffusion in style. In: Proceedings of the IEEE/CVF International Conference on Computer Vision (ICCV). pp. 2251--2261 (October 2023)

\bibitem{frank2020leveraging}
Frank, J., Eisenhofer, T., Sch{\"o}nherr, L., Fischer, A., Kolossa, D., Holz, T.: Leveraging frequency analysis for deep fake image recognition. In: International conference on machine learning. pp. 3247--3258. PMLR (2020)

\bibitem{goodfellow2014generative}
Goodfellow, I., Pouget-Abadie, J., Mirza, M., Xu, B., Warde-Farley, D., Ozair, S., Courville, A., Bengio, Y.: Generative adversarial nets. Advances in neural information processing systems  \textbf{27} (2014)

\bibitem{goodfellow2020generative}
Goodfellow, I., Pouget-Abadie, J., Mirza, M., Xu, B., Warde-Farley, D., Ozair, S., Courville, A., Bengio, Y.: Generative adversarial networks. Communications of the ACM  \textbf{63}(11),  139--144 (2020)

\bibitem{googlefctools2024}
Google: Google fact check tools (2024), \url{https://newsinitiative.withgoogle.com/id/resources/trainings/google-fact-check-tools/}, accessed 4th Jul. 2024

\bibitem{gregor2015draw}
Gregor, K., Danihelka, I., Graves, A., Rezende, D., Wierstra, D.: Draw: A recurrent neural network for image generation. In: International conference on machine learning. pp. 1462--1471. PMLR (2015)

\bibitem{guillaro2023trufor}
Guillaro, F., Cozzolino, D., Sud, A., Dufour, N., Verdoliva, L.: Trufor: Leveraging all-round clues for trustworthy image forgery detection and localization. In: Proceedings of the IEEE/CVF conference on computer vision and pattern recognition. pp. 20606--20615 (2023)

\bibitem{guo2023hierarchical}
Guo, X., Liu, X., Ren, Z., Grosz, S., Masi, I., Liu, X.: Hierarchical fine-grained image forgery detection and localization. In: Proceedings of the IEEE/CVF Conference on Computer Vision and Pattern Recognition. pp. 3155--3165 (2023)

\bibitem{hadwiger2021forchheim}
Hadwiger, B., Riess, C.: The forchheim image database for camera identification in the wild. In: Pattern Recognition. ICPR International Workshops and Challenges: Virtual Event, January 10--15, 2021, Proceedings, Part VI. pp. 500--515. Springer (2021)

\bibitem{he2016deep}
He, K., Zhang, X., Ren, S., Sun, J.: Deep residual learning for image recognition. In: Proceedings of the IEEE conference on computer vision and pattern recognition. pp. 770--778 (2016)

\bibitem{ho2020denoising}
Ho, J., Jain, A., Abbeel, P.: Denoising diffusion probabilistic models. Advances in neural information processing systems  \textbf{33},  6840--6851 (2020)

\bibitem{ju2022fusing}
Ju, Y., Jia, S., Ke, L., Xue, H., Nagano, K., Lyu, S.: Fusing global and local features for generalized ai-synthesized image detection. In: 2022 IEEE International Conference on Image Processing (ICIP). pp. 3465--3469. IEEE (2022)

\bibitem{kang2023scaling}
Kang, M., Zhu, J.Y., Zhang, R., Park, J., Shechtman, E., Paris, S., Park, T.: Scaling up gans for text-to-image synthesis. In: Proceedings of the IEEE/CVF Conference on Computer Vision and Pattern Recognition. pp. 10124--10134 (2023)

\bibitem{karageorgiou2024fusion}
Karageorgiou, D., Kordopatis-Zilos, G., Papadopoulos, S.: Fusion transformer with object mask guidance for image forgery analysis. In: Proceedings of the IEEE/CVF Conference on Computer Vision and Pattern Recognition. pp. 4345--4355 (2024)

\bibitem{karras2018progressive}
Karras, T., Aila, T., Laine, S., Lehtinen, J.: Progressive growing of gans for improved quality, stability, and variation. In: International Conference on Learning Representations (2018)

\bibitem{karras2019style}
Karras, T., Laine, S., Aila, T.: A style-based generator architecture for generative adversarial networks. In: Proceedings of the IEEE/CVF conference on computer vision and pattern recognition. pp. 4401--4410 (2019)

\bibitem{kordopatis2022dns}
Kordopatis-Zilos, G., Tzelepis, C., Papadopoulos, S., Kompatsiaris, I., Patras, I.: Dns: Distill-and-select for efficient and accurate video indexing and retrieval. International Journal of Computer Vision  \textbf{130}(10),  2385--2407 (2022)

\bibitem{koutlis2024leveraging}
Koutlis, C., Papadopoulos, S.: Leveraging representations from intermediate encoder-blocks for synthetic image detection. arXiv preprint arXiv:2402.19091  (2024)

\bibitem{kuznetsova2020open}
Kuznetsova, A., Rom, H., Alldrin, N., Uijlings, J., Krasin, I., Pont-Tuset, J., Kamali, S., Popov, S., Malloci, M., Kolesnikov, A., et~al.: The open images dataset v4: Unified image classification, object detection, and visual relationship detection at scale. International journal of computer vision  \textbf{128}(7),  1956--1981 (2020)

\bibitem{kwon2022learning}
Kwon, M.J., Nam, S.H., Yu, I.J., Lee, H.K., Kim, C.: Learning jpeg compression artifacts for image manipulation detection and localization. International Journal of Computer Vision  \textbf{130}(8),  1875--1895 (2022)

\bibitem{li2024masksim}
Li, Y., Bammey, Q., Gardella, M., Nikoukhah, T., Morel, J.M., Colom, M., Von~Gioi, R.G.: Masksim: Detection of synthetic images by masked spectrum similarity analysis. In: Proceedings of the IEEE/CVF Conference on Computer Vision and Pattern Recognition. pp. 3855--3865 (2024)

\bibitem{lin2024detecting}
Lin, L., Gupta, N., Zhang, Y., Ren, H., Liu, C.H., Ding, F., Wang, X., Li, X., Verdoliva, L., Hu, S.: Detecting multimedia generated by large ai models: A survey. arXiv preprint arXiv:2402.00045  (2024)

\bibitem{lin2014microsoft}
Lin, T.Y., Maire, M., Belongie, S., Hays, J., Perona, P., Ramanan, D., Doll{\'a}r, P., Zitnick, C.L.: Microsoft coco: Common objects in context. In: Computer Vision--ECCV 2014: 13th European Conference, Zurich, Switzerland, September 6-12, 2014, Proceedings, Part V 13. pp. 740--755. Springer (2014)

\bibitem{liu2020global}
Liu, Z., Qi, X., Torr, P.H.: Global texture enhancement for fake face detection in the wild. In: Proceedings of the IEEE/CVF conference on computer vision and pattern recognition. pp. 8060--8069 (2020)

\bibitem{ojha2023towards}
Ojha, U., Li, Y., Lee, Y.J.: Towards universal fake image detectors that generalize across generative models. In: Proceedings of the IEEE/CVF Conference on Computer Vision and Pattern Recognition. pp. 24480--24489 (2023)

\bibitem{park2019semantic}
Park, T., Liu, M.Y., Wang, T.C., Zhu, J.Y.: Semantic image synthesis with spatially-adaptive normalization. In: Proceedings of the IEEE/CVF conference on computer vision and pattern recognition. pp. 2337--2346 (2019)

\bibitem{podell2023sdxl}
Podell, D., English, Z., Lacey, K., Blattmann, A., Dockhorn, T., M{\"u}ller, J., Penna, J., Rombach, R.: Sdxl: Improving latent diffusion models for high-resolution image synthesis. arXiv preprint arXiv:2307.01952  (2023)

\bibitem{radford2021learning}
Radford, A., Kim, J.W., Hallacy, C., Ramesh, A., Goh, G., Agarwal, S., Sastry, G., Askell, A., Mishkin, P., Clark, J., et~al.: Learning transferable visual models from natural language supervision. In: International conference on machine learning. pp. 8748--8763. PMLR (2021)

\bibitem{rombach2022high}
Rombach, R., Blattmann, A., Lorenz, D., Esser, P., Ommer, B.: High-resolution image synthesis with latent diffusion models. In: Proceedings of the IEEE/CVF conference on computer vision and pattern recognition. pp. 10684--10695 (2022)

\bibitem{schinas2024sidbench}
Schinas, M., Papadopoulos, S.: Sidbench: A python framework for reliably assessing synthetic image detection methods. arXiv preprint arXiv:2404.18552  (2024)

\bibitem{sha2023fake}
Sha, Z., Li, Z., Yu, N., Zhang, Y.: De-fake: Detection and attribution of fake images generated by text-to-image generation models. In: Proceedings of the 2023 ACM SIGSAC Conference on Computer and Communications Security. pp. 3418--3432 (2023)

\bibitem{sohl2015deep}
Sohl-Dickstein, J., Weiss, E., Maheswaranathan, N., Ganguli, S.: Deep unsupervised learning using nonequilibrium thermodynamics. In: International conference on machine learning. pp. 2256--2265. PMLR (2015)

\bibitem{tan2024rethinking}
Tan, C., Zhao, Y., Wei, S., Gu, G., Liu, P., Wei, Y.: Rethinking the up-sampling operations in cnn-based generative network for generalizable deepfake detection. In: Proceedings of the IEEE/CVF Conference on Computer Vision and Pattern Recognition. pp. 28130--28139 (2024)

\bibitem{tan2023learning}
Tan, C., Zhao, Y., Wei, S., Gu, G., Wei, Y.: Learning on gradients: Generalized artifacts representation for gan-generated images detection. In: Proceedings of the IEEE/CVF Conference on Computer Vision and Pattern Recognition. pp. 12105--12114 (2023)

\bibitem{wang2020cnn}
Wang, S.Y., Wang, O., Zhang, R., Owens, A., Efros, A.A.: Cnn-generated images are surprisingly easy to spot... for now. In: Proceedings of the IEEE/CVF conference on computer vision and pattern recognition. pp. 8695--8704 (2020)

\bibitem{wang2023dire}
Wang, Z., Bao, J., Zhou, W., Wang, W., Hu, H., Chen, H., Li, H.: Dire for diffusion-generated image detection. In: Proceedings of the IEEE/CVF International Conference on Computer Vision. pp. 22445--22455 (2023)

\bibitem{wang2022diffusiondb}
Wang, Z.J., Montoya, E., Munechika, D., Yang, H., Hoover, B., Chau, D.H.: Diffusiondb: A large-scale prompt gallery dataset for text-to-image generative models. arXiv preprint arXiv:2210.14896  (2022)

\bibitem{zhang2023text}
Zhang, C., Zhang, C., Zhang, M., Kweon, I.S.: Text-to-image diffusion models in generative ai: A survey. arXiv preprint arXiv:2303.07909  (2023)

\bibitem{zhang2023adding}
Zhang, L., Rao, A., Agrawala, M.: Adding conditional control to text-to-image diffusion models. In: Proceedings of the IEEE/CVF International Conference on Computer Vision. pp. 3836--3847 (2023)

\bibitem{zhong2023rich}
Zhong, N., Xu, Y., Qian, Z., Zhang, X.: Patchcraft: Exploring texture patch for efficient ai-generated image detection. arXiv preprint arXiv:2311.12397  (2023)

\bibitem{zhu2017unpaired}
Zhu, J.Y., Park, T., Isola, P., Efros, A.A.: Unpaired image-to-image translation using cycle-consistent adversarial networks. In: Proceedings of the IEEE international conference on computer vision. pp. 2223--2232 (2017)

\bibitem{zhu2023genimage}
Zhu, M., Chen, H., Yan, Q., Huang, X., Lin, G., Li, W., Tu, Z., Hu, H., Hu, J., Wang, Y.: Genimage: A million-scale benchmark for detecting ai-generated image. Advances in Neural Information Processing Systems  \textbf{36} (2023)

\end{thebibliography}
\end{document}